\documentclass[10pt, a4paper]{scrartcl}

\usepackage{latexsym}
\usepackage{amssymb}
\usepackage[nonamelimits]{amsmath}
\usepackage[authoryear, round]{natbib}
\usepackage{url}

\usepackage{color}
\definecolor{myred}{rgb}{0.5,0,0}
\definecolor{myblue}{rgb}{0,0,0.75}
\definecolor{mygreen}{rgb}{0,0.5,0}

\usepackage{ifpdf}

\ifpdf
   \usepackage[pdftex]{graphicx}
   \usepackage[pdftex,
               pdftitle={A plug-in approach to maximising precision at the top},
               pdfsubject={},
               pdfkeywords={Plug-in classifier, precision at the top, precision@K, thresholding},
               pdfauthor={Dirk Tasche},
               pdfstartview=FitR,
               breaklinks=true,
               colorlinks=true,
               citecolor=myred,
               linkcolor=myblue,
               urlcolor=mygreen]{hyperref}
\else
   \usepackage[dvips]{graphicx}
   \usepackage{hyperref}
   \usepackage{rotating}
\fi

\newtheorem{theorem}{Theorem}[section]

\newtheorem{remark}[theorem]{Remark}

\newtheorem{definition}[theorem]{Definition}

\newtheorem{assumption}[theorem]{Assumption}

\newcommand{\qed}{$\Box$}

\numberwithin{equation}{section}

\title{A plug-in approach to maximising precision at the top and recall at the top}

\author{%
Dirk Tasche\thanks{E-mail: dirk.tasche@gmx.net}}

\date{}

\begin{document}

\maketitle

\begin{abstract}
For information retrieval and binary classification, we show that precision at the top (or precision@k) 
and recall at the top (or recall@k) are maximised by thresholding 
the posterior probability of the positive class. This finding is a consequence 
of a result on constrained minimisation of the cost-sensitive expected classification error which
generalises an earlier related result from the literature.
\\[1ex]
\textsc{Keywords:} Plug-in classifier, precision at the top, precision@k, recall at the top, recall@k, thresholding.  
\end{abstract}


\section{Introduction}
\label{se:intro}

Information retrieval and binary classification can be considered equivalent problems in principle. Information retrieval
means to mark documents in a set of candidate documents as relevant or non-relevant for some question, on the basis
of the properties of the documents. For binary classification, the problem is to distinguish between the `positive' and
`negative' instances from a dataset, based on the features of the instances. Hence, from an abstract point of view,
information retrieval is a special case of binary classification, with the documents being instances, the
document properties being features and `relevant' being translated as `positive'. 

In practice, however, the general concepts from binary classification are not always helpful for information retrieval
applications. The fact that often the proportion of relevant documents in a set of documents subject to a search is small
or even very small is only one of the reasons for information retrieval to be considered a field of research for its own.
As a consequence, some performance measures for information retrieval methods differ from those in use 
for binary classifiers or are called by different names.

Precision and recall are possibly the most popular performance measures 
\citep[see Chapter~8 of][for a list of performance measures]{manningInformationRetrieval} for information retrieval methods:
\begin{itemize}
\item Precision is the proportion of documents (instances) that are truly relevant (positive) among those documents
which have been predicted relevant (positive). The term precision is also commonly used (with the same meaning) in 
binary classification.
\item Recall is the proportion of documents (instances) that are predicted relevant (positive) among those documents
which are truly relevant (positive). In binary classification terminology, recall means the same as `true positive rate' or
`sensitivity'.
\end{itemize}

Both precision and recall focus on the performance of a classifier to correctly predict positive instances. This is in contrast
to `accuracy', the most popular performance measure for binary classification which reflects the expected classification error, i.e.\
the probability of positive instances to be predicted negative and of negative instances to  be predicted positive. For information
retrieval in web searches, the focus is not only on the correct prediction of positive instances but also on the 
correct prediction among the top rank predicted positive instances.

`Precision at k' (or `precision at the top')\footnote{%
Like in \citet{kar2015surrogate}, in this note 
`precision at k' or `precision at the top'
are understood to mean the same concept.} and `recall at k` (or `recall at the top')
was mentioned in \citet{Joachims/05a} as an example of a non-linear performance measure 
for the training of classifiers that can be efficiently treated with support vector methods. \citeauthor{Joachims/05a} describes
the reason of why these performance measures are of interest as follows \citep[][Section~4.1]{Joachims/05a}:
``In Web search engines, most
users scan only the first few links that are presented.
Therefore, a common way to evaluate such systems is
to measure precision only on these (e.g.\ ten) positive
predictions. Similarly, in an archival retrieval system
not precision, but recall might be the most indicative
measure. For example, what fraction of the total number
of relevant documents did a user find after scanning
the top 100 documents. Following this intuition,
Prec@k and Rec@k measure the precision and recall of
a classifier that predicts exactly k documents to be
positive.''
The related constrained maximisation problems have been studied in  a number of papers since 
\citep[see][and the references therein]{MackeyLuoEban}.

It is well-known that thresholding the posterior positive class probability
provides an optimal plug-in classifier for Neyman-Pearson classification -- a
similar constrained optimisation problem \citep{tong2013plug}. 
It is less well-known that thresholding the posterior positive class probability also
provides an optimal plug-in classifier for precision at the top and recall at the top. This is a consequence of 
a result by \citet{clemenccon2007ranking} on `classification with a mass constraint'.
 
In this note, we present a generalisation of this result by \citet{clemenccon2007ranking}.
We show that appropriate thresholding of the posterior positive class probability gives
an optimal plug-in classifier for minimising the `expected cost-sensitive error' criterion 
in the presence of a constraint on the predicted positive rate. Our result (Theorem~\ref{th:general} below) 
may be interpreted as a result in between the 
characterisation of globally optimal Bayes classifiers and the characterisation of most powerful tests from the Neyman-Pearson 
lemma. 

More precisely, Theorem~\ref{th:general} is a generalisation of Proposition~1 of \citet{clemenccon2007ranking}
in three ways: 
\begin{itemize}
\item The theorem shows that the RDC (randomized decision classifier) 
version of the optimal classifier $C_{u_0}^\ast$ of \citet{clemenccon2007ranking}
minimises the expected misclassification cost defined by \eqref{eq:cost} below for arbitrary $a, b \ge 0$ with $a+b>0$.
Since Proposition~1 of \citet{clemenccon2007ranking} only deals with minimisation of the error probability, thus Theorem~\ref{th:general} 
is its cost-sensitive generalisation.
\item Theorem~\ref{th:general} also covers the case of discontinuous distributions of the posterior positive class probability. 
\item Theorem~\ref{th:general} refines Proposition~1 of \citet{clemenccon2007ranking} by dealing with more types of constraints
for the predicted positive rate.
\end{itemize}

There are two potential applications of Theorem~\ref{th:general} in practice:
\begin{itemize}
\item For benchmarking of other optimisation algorithms \citep[see e.g.][]{kar2015surrogate}
against an exact solution in a setting where the exact solution can be calculated. The binormal model with equal
variances is an example of such a setting. See Section~\ref{se:binormal} below.
\item Given recent progress in the estimation of posterior probabilities \citep[see][and the references
therein]{kull2017betacalibration}, a plug-in approach based on thresholding the posterior positive 
class probability might turn out to have a competitive edge against other approaches.
\end{itemize}

This note is organised as follows:
\begin{itemize}
\item In Section~\ref{se:second}, the notation needed for precisely stating and proving the generalisation Theorem~\ref{th:general} of the
constrained optimisation problem of \citet{clemenccon2007ranking} is provided and the proof of the theorem is presented. In particular,
the notion of randomised decision classifier (RDC) is introduced. The section concludes with some comments on Theorem~\ref{th:general} and
easy conclusions.
\item Section~\ref{se:precision} shows how Theorem~\ref{th:general} implies an optimal solution to the problem of maximising
precision at the top and recall at the top.
\item In Section~\ref{se:binormal}, we illustrate the optimal plug-in classifier for precision at the top and recall at the top
in the simple binormal setting where the feature distributions are normal with equal variances.
\item Section~\ref{se:concl} concludes the note.
\end{itemize}


\section{Constrained minimisation of cost-sensitive expected error}
\label{se:second}

We discuss binary classification and the properties of classifiers in a probabilistic setting specified 
by a probability space as it was done by many authors before \citep[see, e.g.][]{vanTrees}. The notation
used in this paper is broadly aligned with the notation specified in Section~1~C of \citet{scott2005neyman}.
Like \citet{scott2005neyman}, \citet{clemenccon2007ranking}, \citet{koyejo2014consistent} and other machine learning 
researchers dealing with the theoretical results we use the language of measure theory in order to be able to precisely
state our results.

Accordingly, the probability space $(\Omega, \mathcal{A}, \mathrm{P})$ describes the experiment of choosing
an instance at random. 
The instance has a class label and features. The features can be observed immediately while,
depending on whether the probability space is interpreted as a training sample or target sample (sometimes
also called test sample), 
the label is also observable at once or can be observed only with some delay.
We interpret $\mathcal{A}$ as the $\sigma$-field \citep[see, e.g.][Section~2]{billingsley3rd}
of all admissible events,
including events that cannot yet be observed. In addition, we have a $\sigma$-field $\mathcal{H}$
which is the family of the events that can be observed now.
The event $A$ with $A\in\mathcal{A}$ but $A\notin\mathcal{H}$ reveals the instance's class label. 
If $A$ occurs the instance has got class label $1$ (positive). 
If $A^c = \Omega\backslash{}A$ occurs the instance's label is $-1$ (negative). 

\begin{assumption}\label{as:0}\ 
\begin{itemize}
\item $(\Omega, \mathcal{A}, \mathrm{P})$ is a probability space\footnote{%
See text books on probability theory like \citet{Durrett} or \citet{billingsley3rd} for the formal definition.}. 
This space describes the experiment
	of selecting an instance from a population at random and observing its features and (typically with some
	delay) class label.
\item $A \in \mathcal{A}$ is a fixed event with $0<\mathrm{P}[A]<1$. If $A$ is observed, the instance's 
	class label is 1, otherwise if $A^c = \Omega\backslash{}A$ is observed, the instance's class label is -1. 
\item $\mathcal{H} \subset \mathcal{A}$ is a sub-$\sigma$-field of
$\mathcal{A}$ such that $A \notin \mathcal{H}$. $\mathcal{H}$ is the $\sigma$-field of immediately observable
events and, in particular, features.
\end{itemize}
\end{assumption}

In a binary classification problem setting, typically there are random variables $X: \Omega \to \mathcal{X} \subset\mathbb{R}^d$
for some $d \in \mathbb{N}$ (vector of explanatory variables or features) 
and $Y:\Omega \to \{-1, 1\}$ (dependent or class variable) such that $\mathcal{H} = \sigma(X)$ and
$Y^{-1}(\{1\}) = A$.

In the machine learning literature, a classifier is function that maps an observed feature vector to 1 or $-1$. Classifiers
are interpreted as predictors of the class (positive or negative) of an instance on the basis of the instance's features.
In our setting, a classifier is an $\mathcal{H}$-measurable random variable $H$ with values in the set $\{-1, 1\}$. 
The $\mathcal{H}$-measurability of $H$ reflects the fact that the value of the classifier depends only on the features
of the instance in question because the instance's class is assumed to be unknown at the time the classifier is applied.

For the purpose of this note, we make use of a more general than usual definition of classifier in order to be able to
describe the main result in the most rigorous manner. This concept of classifier is called randomized decision classifier.
It is the equivalent of randomized tests which have been mentioned in the machine learning literature in the 
context of Neyman-Pearson classification \citep{scott2005neyman, tong2013plug}.

\begin{definition}[Randomized decision classifier (RDC)]\label{de:classifier}
Under Assumption~\ref{as:0}, a \emph{randomized decision classifier (RDC)} is an $\mathcal{H}$-measurable random
variable with values in the unit interval $[0,1]$. A (deterministic or ordinary) 
\emph{classifier (OC)} is an $\mathcal{H}$-measurable random
variable with values in the set $\{0,1\}$. For an RDC $H$, its expected value $\mathrm{E}[H]$ is called
\emph{predicted positive rate}.
\end{definition}
In the following, $H$ typically denotes an RDC or OC in the sense of Definition~\ref{de:classifier}. 
Note that each OC is also an RDC.  A classifier $H$ (OC or RDC) is used in two steps to predict the class of an instance:
\begin{itemize}
\item In the first step, depending on the features of the instance, the value of $H$ is determined.
\item The second step is to perform an independent random experiment which gives `positive' with probability $H$ and 
`negative' with probability $1-H$. The outcome of this experiment is the prediction of the instance's class.
\end{itemize}
In the case where $H$ is an OC (i.e.\ takes on only values 0 or 1), the second step is redundant in the sense that no
experiment needs to be performed since $H =1$ implies `prediction is positive' while $H=0$ means `prediction is negative'.

By Definition~\ref{de:classifier}, an RDC $H$ is a $\mathcal{H}$-measurable random variable $\Omega \to [0,1]$, in 
analogy to the concept of randomized test from the Neyman-Pearson lemma. This is different
to the `randomized classifier' notion in the machine learning literature \citep[see][and the references
therein]{pmlr-v76-thiemann17a}. There a `randomized classifier' is 
a random draw from a set of ordinary classifiers. Note, however, that there are two interpretations of $H$ in the
RDC sense: 
\begin{enumerate}
\item $H$ is the probability of prediction `positive' in an additional independent experiment (the randomized
decision).
\item Each time before $H$ is applied, for each $x \in \mathcal{X}$ 
a 0 or 1 decision is made at random with probability $H(x)$ for 1. This multitude of random experiments generates 
a random selection from the set of all ordinary classifiers $\{-1,1\}^\mathcal{X}$ that is used to predict the instance's class.
\end{enumerate}
The second interpretation of RDC shows that RDCs can be considered special cases of 
randomized classifiers.  

At first glance, the concept of RDC might seem rather unintuitive. To evaluate a classifier with possibly large numerical effort and
then decide by chance is not a convincing approach. However it turns out below in Theorem~\ref{th:general} that there
is a best RDC that is `nearly' -- in a sense that is specified below in Remark~\ref{rm:comments}~(iii) -- 
deterministic. Thus we buy mathematical perfection in 
the sense of a result `without gaps' (it holds also for the case of non-unique quantiles and discontinuous distributions) 
at the price of a most of the time negligible deviation from deterministic classifiers. This is a time-honoured approach that
was applied before to statistical test theory and Neyman-Pearson classification \citep{scott2005neyman, tong2013plug}.

Define the \emph{expected misclassification cost} $L_{a,b}(H)$ for an RDC $H$ and fixed $a, b \ge 0$ with $a+b>0$ by\footnote{%
$\mathbf{1}_S$ denotes the indicator function of the set $S$, i.e.\ $\mathbf{1}_S(s)=1$ for $s\in S$ and
$\mathbf{1}_S(s)=0$ for $s\notin S$.}
\begin{equation}\label{eq:cost}
	L_{a,b}(H) \ = \ a\,\mathrm{E}[(1-H)\,\mathbf{1}_A] + b\,\mathrm{E}[H\,\mathbf{1}_{A^c}].
\end{equation}
In \eqref{eq:cost}, 
\begin{align*}
\mathrm{E}[(1-H)\,\mathbf{1}_A] &\ =\ \mathrm{P}[\text{Class is positive and $H$ predicts negative}]\\
\intertext{and}
\mathrm{E}[H\,\mathbf{1}_{A^c}] &\ =\ \mathrm{P}[\text{Class is negative and $H$ predicts positive}]
\end{align*}
are the 
probabilities of the two possible errors resulting from the application of $H$. In the language of test theory
$\mathrm{E}[(1-H)\,\mathbf{1}_A]$ is the probability of a \emph{type I error} while $\mathrm{E}[H\,\mathbf{1}_{A^c}]$
is the probability of a \emph{type II error}. Thus $L_{a,b}(H)$ is a cost-weighted average of the two error
probabilities.

For the statement of Theorem~\ref{th:general} the notion of quantile is crucial. We use a definition which takes
account of the fact that sometimes there is more than one choice for the quantile of a distribution at a certain 
level.
\begin{definition}\label{de:quant}
Let $Z$ be a real-valued random variable and  $\alpha \in (0,1)$ be fixed. Then each $z\in\mathbb{R}$ with
\begin{equation}\label{eq:quant}
P[Z < z]\ \le\ \alpha\ \le\ P[Z \le z]
\end{equation}
 is an $\alpha$-quantile of $Z$ (and of the distribution of $Z$). 
\end{definition}
Note that the set of $\alpha$-quantiles is non-empty for all $\alpha \in (0,1)$ and either has exactly one element
or is a closed interval. In the literature and in practice, often `$\alpha$-quantile' is understood as 
$\min \{z: P[Z \le z] \ge \alpha\}$, the lower limit of that interval.

Recall the notion of probability of an event conditional on a $\sigma$-field as 
defined in standard text books on probability theory \citep[e.g.][Section~33]{billingsley3rd}. In the context of 
Assumption~\ref{as:0}, $\mathrm{P}[A\,|\,\mathcal{H}]$ denotes the probability 
of $A$ conditional on $\mathcal{H}$ (`posterior probability' in machine learning terminology).  
$\mathrm{P}[A\,|\,\mathcal{H}]$ can be characterised as 
$\mathcal{H}$-measurable random variable such that
\begin{gather*}
0\ \le \ \mathrm{P}[A\,|\,\mathcal{H}]\ \le 1,\\
\text{and}\quad\mathrm{E}\bigl[\mathrm{P}[A\,|\,\mathcal{H}]\,Z\bigr]\ = \ \mathrm{E}[\mathbf{1}_A\,Z]
\end{gather*}
for all bounded $\mathcal{H}$-measurable random variables $Z$.

For the proof of the following Theorem~\ref{th:general}, we revisit the proof of Proposition~1 of \citet{clemenccon2007ranking} 
and the classical proof of optimality of
the Bayes classifier for the cost-sensitive error criterion as given in Section~2.2.1 of \citet{vanTrees} or in
Section~1.3 of \citet{Elkan01}.

\begin{theorem}\label{th:general} Under Assumption~\ref{as:0}, 
let $a, b \ge 0$ with $a+b>0$ be fixed. Define $L_{a,b}(H)$  by 
\eqref{eq:cost}. Let $0 < \alpha <1$ and any $(1-\alpha)$-quantile $q$ of the posterior class probability
$\mathrm{P}[A\,|\,\mathcal{H}]$ be fixed. Define $H_q$ by
\begin{equation}\label{eq:Hq}
H_q \ = \ \begin{cases}\mathbf{1}_{\{\mathrm{P}[A\,|\,\mathcal{H}] > q\}} + \frac{\alpha - \mathrm{P}\bigl[\mathrm{P}[A\,|\,\mathcal{H}] > q\bigr]}
        {\mathrm{P}\bigl[\mathrm{P}[A\,|\,\mathcal{H}] = q\bigr]}\,\mathbf{1}_{\{\mathrm{P}[A\,|\,\mathcal{H}] = q\}}, & \text{if}\ \/
        \mathrm{P}\bigl[\mathrm{P}[A\,|\,\mathcal{H}] = q\bigr] > 0,\\
        \mathbf{1}_{\{\mathrm{P}[A\,|\,\mathcal{H}] > q\}}, & \text{if}\ \/ 
        \mathrm{P}\bigl[\mathrm{P}[A\,|\,\mathcal{H}] = q\bigr] = 0.
        \end{cases}
\end{equation}
Then $H_q$ is an RDC in the sense of Definition~\ref{de:classifier} with $\mathrm{E}[H] = \alpha$ such that
the following three statements hold:
\begin{itemize}
\item[(i)] $q < \frac b{a+b}$ \quad$\Rightarrow$ \quad
$H_q = \arg \underset{H\, \mathrm{is\,RDC},\,\mathrm{E}[H] \ge \alpha}{\min} L_{a,b}(H)$.
\item[(ii)] $q > \frac b{a+b}$ \quad$\Rightarrow$ \quad
$H_q = \arg \underset{H\, \mathrm{is\,RDC},\,\mathrm{E}[H] \le \alpha}{\min} L_{a,b}(H)$.
\item[(iii)] $q = \frac b{a+b}$ \quad$\Rightarrow$ \quad
$H_q = \arg \underset{H\, \mathrm{is\,RDC}}{\min} L_{a,b}(H)$.
\end{itemize}
\end{theorem}
Theorem~\ref{th:general} may be read in two ways:
\begin{enumerate}
\item Fix $\alpha$, calculate $q$ and then select the one of the three statements that
applies.
\item Fix $q$, select the one of the three statements that
applies and then determine $\alpha$.
\end{enumerate}
In Section~\ref{se:binormal} below, we provide an example of how the calculation of $q$ might look like in
practice.

The classical result on the optimal cost-sensitive Bayes classifier (see Section~2.2.1 of \citeauthor{vanTrees}, 
\citeyear{vanTrees}, or Section~1.3 of \citeauthor{Elkan01}, \citeyear{Elkan01}) can be phrased as follows
in the notation of this note:
\begin{equation}\label{eq:classical}
\mathbf{1}_{\{\mathrm{P}[A\,|\,\mathcal{H}] > \frac b{a+b}\}}\ =\ \arg \underset{H\, \mathrm{is\,OC}}{\min} L_{a,b}(H).
\end{equation}
Hence, in the case of $\mathrm{P}\bigl[\mathrm{P}[A\,|\,\mathcal{H}] = \frac b{a+b}\bigr] > 0$, the optimal
classifiers according to Theorem~\ref{th:general} and according to the classical result differ. 
Nonetheless, in both cases the otimal value of $L_{a,b}(H)$ is the same. See Remark~\ref{rm:comments} (i) below for
more detail on this observation.

\textbf{Proof of Theorem~\ref{th:general}.} Let any RDC $H$ in the sense of Definition~\ref{de:classifier} be given. Define 
$H_q$ by \eqref{eq:Hq}. Observe that then $0 \le H_q \le 1$ and $H_q$ is an RDC with 
$\mathrm{E}[H_q] = \alpha$. With some algebra, it can be shown that 
\begin{align}
L_{a,b}(H) & =  a\,\mathrm{P}[A] + \bigl(b-(a+b)\,q\bigr)\,\mathrm{E}[H] + 
	(a+b)\,\mathrm{E}\bigl[(q-\mathrm{P}[A\,|\,\mathcal{H}])\,H\, H_q\bigr]\notag \\
	& \qquad + (a+b)\,\mathrm{E}\bigl[(q-\mathrm{P}[A\,|\,\mathcal{H}])\,
			H\,(1-H_q)\bigr]\notag\\
			& \ge a\,\mathrm{P}[A] + \bigl(b-(a+b)\,q\bigr)\,\mathrm{E}[H] + 
	(a+b)\,\mathrm{E}\bigl[(q-\mathrm{P}[A\,|\,\mathcal{H}])\,H\,H_q\bigr]\notag\\
& \ge a\,\mathrm{P}[A] + \bigl(b-(a+b)\,q\bigr)\,\mathrm{E}[H] + 
	(a+b)\,\mathrm{E}\bigl[(q-\mathrm{P}[A\,|\,\mathcal{H}])\,H_q\bigr].\label{eq:end}
\end{align}
In case $q < \frac b{a+b}$ we have $b-(a+b)\,q > 0$. Then  it holds that
$\bigl(b-(a+b)\,q\bigr)\,\mathrm{E}[H] \ge \bigl(b-(a+b)\,q\bigr)\,\mathrm{E}[H_q]$ for $\mathrm{E}[H] \ge \alpha$.
By \eqref{eq:end}, this implies (i).

In case $q > \frac b{a+b}$ we have $b-(a+b)\,q < 0$. Then  it holds that
$\bigl(b-(a+b)\,q\bigr)\,\mathrm{E}[H] \ge \bigl(b-(a+b)\,q\bigr)\,\mathrm{E}[H_q]$ for $\mathrm{E}[H] \le \alpha$.
From this observation and \eqref{eq:end}, statement (ii) follows.
 
In case $q = \frac b{a+b}$ we have $b-(a+b)\,q = 0$. This implies
\begin{equation}\label{eq:equal}
 L_{a,b}(H) \ge 
a\,\mathrm{P}[A]  + 
	(a+b)\,\mathrm{E}\bigl[(q-\mathrm{P}[A\,|\,\mathcal{H}])\,H_q\bigr] = L_{a,b}(H_q)
 \end{equation}  
	for all $H$ without any restriction for $\mathrm{E}[H]$ and hence (iii).
\hfill \qed

Theorem~\ref{th:general} is about 'locally' optimal classifiers in the sense that only classifiers
with the same predicted positive rate are compared. We state this observation more
precisely in item~(iii) of the following remark:

\begin{remark}\label{rm:comments}\ 
\emph{%
\begin{itemize}
\item[(i)] The statement of Theorem~\ref{th:general}~(iii) is similar but not identical to
the classical result \eqref{eq:classical} on the optimal Bayes classifier for the cost-sensitive error criterion.
The difference is the second term in the definition of the optimal RDC $H_q$ as shown in \eqref{eq:Hq} which involves the
factor $\frac{\alpha - \mathrm{P}\bigl[\mathrm{P}[A\,|\,\mathcal{H}] > q\bigr]}
        {\mathrm{P}\bigl[\mathrm{P}[A\,|\,\mathcal{H}] = q\bigr]}$. Actually, close inspection of \eqref{eq:equal} 
in the proof of Theorem~\ref{th:general}~(iii) 
reveals a result  slightly more general than the classical result, namely
\begin{equation}
q = \frac b{a+b} \quad \Rightarrow \quad
H(m) = \arg \underset{H\, \mathrm{is\,RDC}}{\min} L_{a,b}(H),
 \end{equation} 
for all $H(m) = \mathbf{1}_{\{\mathrm{P}[A\,|\,\mathcal{H}] > q\}} + m\,\mathbf{1}_{\{\mathrm{P}[A\,|\,\mathcal{H}] = q\}}$, 
where $0 \le m \le 1$. Observe, however, that if $\mathrm{P}\bigl[\mathrm{P}[A\,|\,\mathcal{H}] = q\bigr]>0$ and 
$\alpha \in (0,1)$ is such that $q$ is an $(1-\alpha)$-quantile, then $m = \frac{\alpha - \mathrm{P}\bigl[\mathrm{P}[A\,|\,\mathcal{H}] > q\bigr]}
        {\mathrm{P}\bigl[\mathrm{P}[A\,|\,\mathcal{H}] = q\bigr]}$ is the only value of $m$ with the property
        $\mathrm{E}[H(m)] = \alpha$.
\item[(ii)] Observe that the RDC defined by \eqref{eq:Hq} is sandwiched by two OCs as defined in item (i):
$$H(0) \ \le \ H_q \ \le H(1).$$
With $H_q$, randomized decisions only have to made in the event $\{\mathrm{P}[A\,|\,\mathcal{H}] = q\}$ whose probability in practice 
tends to be zero or small. Often $H_q$ will be well approximated by both $H(0)$ and $H(1)$ such that there is no need to
take recourse to randomized decisions.
\item[(iii)]
For fixed $\alpha \in (0,1)$, irrespectively of the relation between $q$ and $\frac{b}{a+b}$, Theorem~\ref{th:general} 
implies that for all\/
$a, b \ge 0$ with $a+b > 0$ and $(1-\alpha)$-quantiles $q$ of $\mathrm{P}[A\,|\,\mathcal{H}]$ it holds that
\begin{equation*}
H_q \ =\ \arg \underset{H\,\mathrm{is\,RDC},\,\mathrm{E}[H] = \alpha}{\min} L_{a,b}(H).
\end{equation*}
\item[(iv)] In the case $a = 0$, $b = (1-\mathrm{P}[A])^{-1}$, Theorem~\ref{th:general}~(i) implies for 
$\alpha \in (0,1)$ and any $(1-\alpha)$-quantile $q$ that it holds that
\begin{equation*}\label{eq:fpr}
	H_q \ =\ \arg \underset{H\,\textrm{is\,RDC},\,\mathrm{E}[H] \ge \alpha}{\min} 
		\frac{\mathrm{P}[H\,\mathbf{1}_{A^c}]}{\mathrm{P}[A^c]} \ = \ 
		\arg \underset{H\,\textrm{is\,RDC},\,\mathrm{E}[H] \ge \alpha}{\min} \mathrm{E}[H \,|\, A^c].
\end{equation*}
In the case where $H$ is an OC, we have $\mathrm{E}[H \,|\, A^c] = \mathrm{P}[H = 1\,|\, A^c]$. In any case,
$\mathrm{E}[H \,|\, A^c]$ is called `false positive rate' (FPR). 
\item[(v)] In the case $a = \mathrm{P}[A]^{-1}$ and $b = 0$, Theorem~\ref{th:general}~(ii) implies for 
$\alpha \in (0,1)$ and any $(1-\alpha)$-quantile $q$ that it holds that
\begin{equation*}\label{eq:tpr}
	H_q \ =\ \arg \underset{H\,\textrm{is\,RDC},\,\mathrm{E}[H] \le \alpha}{\max} 
		\frac{\mathrm{E}[H \,\mathbf{1}_A]}{\mathrm{P}[A]} \ = \ 
		\arg \underset{H\,\textrm{is\,RDC},\,\mathrm{E}[H] \le \alpha}{\max} \mathrm{E}[H \,|\, A].
\end{equation*}
In the case where $H$ is an OC, we have $\mathrm{E}[H \,|\, A] = \mathrm{P}[H =1\,|\, A]$. In any case, 
$\mathrm{E}[H \,|\, A]$ is called `true positive rate' (TPR) or
`recall'.
\end{itemize}%
} 
\end{remark}
 
\section{Application to precision at the top and recall at the top}
\label{se:precision}

\citet[][Section~1]{BoydAccuracyAtTheTop} observe that ``the notion of top k does not generalize
to new data. For what k should one train if the test data in some instances is half the size and
in other cases twice the size? In fact, no generalization guarantee is available for such precision@k
optimization or algorithm.'' \citeauthor{BoydAccuracyAtTheTop} therefore suggest that ``a more 
principled approach in all the applications already mentioned consists of designing algorithms
that optimize accuracy in some top fraction of the scores returned by a real-valued hypothesis.''

Both \citet[][Section~2]{clemenccon2007ranking} and this note in Section~\ref{se:second} follow 
this approach, by choosing the posterior positive class probability as the score and the
range of the posterior positive class probability beyond an appropriately selected quantile as
the `top fraction'.

In the notation of Section~\ref{se:second}, we denote the top fraction by fixed $0 < \alpha < 1$. If $H$ denotes
an RDC or OC in the sense of Definition~\ref{de:classifier} then it follows that under Assumption~\ref{as:0} precision and recall respectively
of $H$ are given by the following equations
\begin{subequations}
\begin{align}
\text{precision}(H) & = \frac{\mathrm{E}[H\,\mathbf{1}_A]}{\mathrm{E}[H]}, \label{eq:prec}\\
\text{recall}(H) & = \frac{\mathrm{E}[H \,\mathbf{1}_A]}{\mathrm{P}[A]}.\label{eq:recall}
\end{align}
\end{subequations}
Maximising recall at the top fraction $\alpha$ (of the score values) then means to solve
this optimisation problem:
\begin{subequations}
\begin{equation}\label{eq:recallProbl}
\underset{H\, \mathrm{is\,RDC}}{\max} \text{recall}(H), \qquad \text{subject to}\quad \mathrm{E}[H] = \alpha.
\end{equation}
Since the optimising RDC $H_q$, as identified in Theorem~\ref{th:general}, satisfies $\mathrm{E}[H_q] = \alpha$,
Remark~\ref{rm:comments}~(v) shows that the RDC $H_q$ given by \eqref{eq:Hq} for any $(1-\alpha)$-quantile $q$
of the posterior class probability $\mathrm{P}[A\,|\,\mathcal{H}]$ solves problem \eqref{eq:recallProbl}:
\begin{equation}\label{eq:recallSol}
H_q \ =\ \arg \underset{H\,\textrm{is\,RDC},\, \mathrm{E}[H] = \alpha}{\max}  \text{recall}(H).
\end{equation}
\end{subequations}
Result \eqref{eq:recallSol} also follows from Proposition~1 of \citet{clemenccon2007ranking}  
because the classification error in Proposition~1 of \citeauthor{clemenccon2007ranking} is minimised when the true positive rate
(i.e.\ recall) is maximised.
\citeauthor{clemenccon2007ranking} call the problem `classification with a mass contraint'. Mass of 
an ordinary classifier in the sense of \citeauthor{clemenccon2007ranking} is $\mathrm{E}[H]$ for an RDC in the context of this
note.

Maximising recall at the top fraction $\alpha$ (of the score values) in the notation of this note means to solve
this optimisation problem:
\begin{equation}\label{eq:precProbl}
\underset{H\, \mathrm{is\,RDC}}{\max} \text{precision}(H), \qquad \text{subject to}\quad \mathrm{E}[H] = \alpha.
\end{equation}
By Definition~\eqref{eq:prec} of $\text{precision}(H)$, this maximisation problem can be equivalently written as
\begin{equation*}
\tfrac{\mathrm{P}[A]}{\alpha}\,\underset{H\, \mathrm{is\,RDC}}{\max} \text{recall}(H), \qquad \text{subject to}\quad \mathrm{E}[H] = \alpha.
\end{equation*}
Hence problems \eqref{eq:precProbl} and \eqref{eq:recallProbl} basically are the same and the RDC $H_q$ from 
\eqref{eq:recallSol} also provides a solution to \eqref{eq:precProbl}. In the following Section~\ref{se:binormal}, a numerical example
of how $H_q$ looks like is given.

 
\section{The binormal case with equal variances}
\label{se:binormal}

We revisit the `binormal model' with equal variances as 
an example that fits into the setting of Assumption~\ref{as:0}. Like in \citet{tasche2017fisher}, benchmarking
a new classifier for maximising precision at the top against the known optimal plug-in classifier in this
simple example may serve as a first test for the usefulness of the candidate classifier.
\begin{itemize}
\item We define $\Omega = \mathbb{R} \times \{-1, 1\}$. On $\mathbb{R}$ and $\{-1, 1\}$, we consider the
    the Borel-$\sigma$-field $\mathcal{B}(\mathbb{R})$ and the power set $\mathcal{P}(\{-1, 1\})$ respectively
    as the relevant sets of observable events.
\item On $\Omega$, we define the projections $X$ and $Y$, i.e.\ for $\omega = (x, y) \in \Omega$ we let
    $X(\omega) = x$ and $Y(\omega) = y$.
\item The $\sigma$-field $\mathcal{A}$ on $\Omega$ is given as the intersection of all $\sigma$-fields on $\Omega$ such that
both projections $X$ and $Y$ are measurable as mappings from $(\Omega, \mathcal{A})$ to $(\mathbb{R}, \mathcal{B}(\mathbb{R}))$ 
and $(\{-1, 1\}, \mathcal{P}(\{-1, 1\}))$ respectively, i.e.\ $\mathcal{A} = \sigma(X,Y)$.
\item With $\mathcal{H} = \sigma(X)$, we have $A = \{Y=1\} \notin \mathcal{A}$ by construction.
\item $\mathrm{P}$ is defined by specifying the marginal distribution of $Y$ with $\mathrm{P}[A] = p \in (0,1)$, 
     and defining the conditional distribution of $X$ given $Y$ as combination of two normal distributions 
     with equal variances:
\begin{subequations} 
\begin{equation}\label{eq:CondNormal}
\begin{split}
    \mathrm{P}[X \in \cdot\,|\,A] & = \mathcal{N}(\nu, \sigma^2),\\
    \mathrm{P}[X \in \cdot\,|\,A^c] & = \mathcal{N}(\mu, \sigma^2). 
\end{split}
\end{equation}
    In \eqref{eq:CondNormal}, we assume that $\mu < \nu$ and $\sigma > 0$. \eqref{eq:CondNormal} implies that the
    distribution of $X$ is given by a mixture of normal distributions\footnote{%
    $\Phi$ denotes the standard normal distribution function $\Phi(x) = \frac{1}{\sqrt{2\,\pi}} \int_{-\infty}^x 
    e^{-\,y^2/2}\,d y$.}
\begin{equation}\label{eq:mix}
\mathrm{P}[X \le x] \ =\ p\,\Phi\left(\frac{x-\nu}{\sigma}\right) +
    (1-p)\,\Phi\left(\frac{x-\mu}{\sigma}\right), \quad x \in \mathbb{R}.
\end{equation}
\end{subequations}
\item The posterior probability $\mathrm{P}[A\,|\,\mathcal{H}]$ in this setting is given by
\begin{equation}\label{eq:post}
    \mathrm{P}[A\,|\,\mathcal{H}] \  = \ \frac{1}{1 + \exp(a\,X + b)},
\end{equation}
with $a = \frac{\mu-\nu}{\sigma^2} < 0$ and $b = \frac{\nu^2-\mu^2}{2\,\sigma^2} + \log\left(\frac{1-p}{p}\right)$.
\end{itemize}
Fix $0 < \alpha < 1$. To determine the solution $H_q$ of (say) \eqref{eq:recallProbl}, by \eqref{eq:post} we need to solve
the following equation for $q$:
\begin{equation*}
\mathrm{P}\left[\tfrac{1}{1 + \exp(a\,X + b)} > q\right]\ = \ 1-\alpha.
\end{equation*}
With a little algebra, we find that $q$ is uniquely determined by
\begin{subequations}
\begin{equation}\label{eq:exp}
q \ = \ \tfrac{1}{1 + \exp(a\,x_\alpha + b)},
\end{equation}
where $x_q$ is the unique solution of
\begin{equation}\label{eq:numeric}
\begin{split}
\alpha & \ = \ \mathrm{P}[X \le x_\alpha]\\
    & \ =\ p\,\Phi\left(\frac{x_\alpha-\nu}{\sigma}\right) +
    (1-p)\,\Phi\left(\frac{x_\alpha-\mu}{\sigma}\right).
\end{split}
\end{equation}
\end{subequations}
From \eqref{eq:mix} and \eqref{eq:exp}, it follows that $\mathrm{P}\bigl[\mathrm{P}[A\,|\,\mathcal{H}]=q\bigr] = 0$.
Hence $H_q$ is not a proper RDC but rather a deterministic ordinary classifier.

There is no closed-form solution $x_\alpha$ for \eqref{eq:numeric}. $x_\alpha$ has to be calculated
by numerical methods. However, applications of the optimal classifier in a population model like the
binormal model described above are not very common. More common are applications based on real-world 
datasets or samples where the problem of determining $H_q$ becomes a problem of quantile estimation.
\citet[][Section~4]{MackeyLuoEban} give an example of a possible approach to this estimation problem.


\section{Conclusions}
\label{se:concl}

Plug-in classifiers in some situations are attractive because they are trained (or estimated) only once and then readily 
adapted to changed circumstances by modifying a threshold. Different constraints to the training criterion like
the k in the k top ranks considered for precision or recall at k provide important examples of such changes of
circumstances. 
In binary classification, the posterior probability of the positive class is a primary candidate
to serve as a plug-in classifier. This is well-known for Neyman-Pearson classification but seems to be
less clear for the problem of finding classifiers  optimal for precision at the top or recall at the top.

In this note, we have shown that indeed the posterior positive class probability, with appropriately chosen thresholds,
maximises precision at the top and recall at the top and hence can be used as a plug-in classifier for
related binary classification and information retrieval problems.
Thanks to recent progress in the estimation of posterior probabilities, a plug-in approach based on thresholding the posterior positive 
class probability appears promising and competitive.



\addcontentsline{toc}{section}{References}

%

\end{document}